\pdfoutput=1
\documentclass[11pt]{article}

\usepackage[]{ACL2023}

\usepackage{times}
\usepackage{latexsym}

\usepackage[T1]{fontenc}
\usepackage[utf8]{inputenc}

\usepackage{microtype}

\usepackage{inconsolata}

\usepackage{graphicx}
\usepackage{booktabs}

\title{Transfer Learning for Text Diffusion Models}

\author{Kehang Han$^{1}$\Thanks{~Equal contribution.} , Kathleen Kenealy$^{1}$\footnotemark[1] , Aditya Barua$^{2}$\footnotemark[1] , Noah Fiedel$^{1}$, Noah Constant$^{1}$ \\
$^{1}$Google DeepMind \quad $^{2}$Google \\
\texttt{\{kehanghan,\,kkenealy,\,adityabarua,\,nfiedel,\,nconstant\}@google.com}
}

\begin{document}
\maketitle
\begin{abstract}

In this report, we explore the potential for \emph{text diffusion} to replace autoregressive (AR) decoding for the training and deployment of large language models (LLMs). We are particularly interested to see whether pretrained AR models can be transformed into text diffusion models through a lightweight adaptation procedure we call ``AR2Diff''. We begin by establishing a strong baseline setup for training text diffusion models. Comparing across multiple architectures and pretraining objectives, we find that training a decoder-only model with a prefix LM objective is best or near-best across several tasks. Building on this finding, we test various transfer learning setups for text diffusion models. On machine translation, we find that text diffusion underperforms the standard AR approach. However, on code synthesis and extractive QA, we find diffusion models trained from scratch outperform AR models in many cases. We also observe quality gains from AR2Diff---adapting AR models to use diffusion decoding. These results are promising given that text diffusion is relatively underexplored and can be significantly faster than AR decoding for long text generation.

\end{abstract}

\section{Introduction}

In recent years, large language models (LLMs) have grown in scale, capability, and popularity \cite{brown2020language, chowdhery2022palm}, and are increasingly used to generate long-form text such as summaries, blocks of code, or in-depth explanations \cite{openai2023gpt4, anil2023palm}. To our knowledge, all popular LLMs are \emph{autoregressive} (AR)---generating one token at a time in textual order, each conditioned on the sequence generated so far. While AR generation is well understood and has been highly optimized, its strict left-to-right factorization may be overly constraining. Generating token-by-token is inherently inefficient, particularly on long but predictable spans of text (e.g., copying a serial number from the context one digit at a time). Additionally, this strict order may not provide the ideal scaffold for planning a composition. Human writers typically outline, draft, revise, and proofread their work, and it seems plausible that machines could benefit from a similar iterative approach.\footnote{``Chain of thought'' prompting \cite{wei2022chain} provides a mechanism for models to reason about or draft the desired output before producing it. However, the final output is still generated autoregressively.}

As an alternative, many \emph{non-AR} decoding methods have been proposed (see section~\S\ref{sec:related_work}), which generate multiple sequence positions in parallel, or make progressive edits to a ``rough'' initial generation. Several of these have shown promising results on specific tasks. For example, SUNDAE's \emph{text diffusion} approach \cite{savinov2022stepunrolled} achieves similar quality to an AR baseline on machine translation while decoding over 2$\times$ faster.

However, despite positive findings, non-AR techniques have failed to gain traction, and remain unused in the space of large language models. We suspect this may be due to the inertia behind classic AR methods, and the high cost and risk of tuning and training large models from scratch using non-standard training losses and decoding methods.

With an eye to lowering this cost of entry and easing the transition to more efficient text generation at scale, in this paper we investigate the potential for adapting existing pretrained AR model checkpoints to perform non-AR generation. We use a simplified version of SUNDAE text diffusion as our canonical non-AR implementation; thus we refer to this lightweight adaptation process as \textbf{AR2Diff (AR to Diffusion)}.

More specifically, we are interested in testing the ability of text diffusion methods to compete at scale in the popular transfer learning setting, where a model is pretrained on unsupervised data and applied to diverse downstream tasks. We conduct a series of experiments comparing text diffusion to AR baselines across different model architectures, tasks, and transfer learning settings.

Our main contributions are: (1) showing that language models pretrained and fine-tuned using text diffusion can be competitive with autoregressive models on several downstream tasks, (2) showing that pretrained AR models can be transformed into diffusion models via a lightweight adaptation.

\section{Related Work}
\label{sec:related_work}

Previous work has explored a wide range of non-autoregressive methods for text generation \cite{gu2018nonautoregressive, lee-etal-2018-deterministic, stern2019insertion, ghazvininejad-etal-2019-mask}.  In the last few years, diffusion models \cite{sohldickstein2015deep} have emerged as the primary technique for \emph{image} generation \cite{rombach2021high, ramesh2022hierarchical, saharia2022photorealistic}. Many recent efforts have applied diffusion methods to \emph{text} generation \cite{savinov2022stepunrolled, li2022diffusionlm, reid2023diffuser, chen2023analog, strudel2022selfconditioned, dieleman2022continuous, zheng2023reparameterized, lin2023text, gong2023diffuseq, yuan2023seqdiffuseq, wu2023ardiffusion}, but none has yet gained adoption in the space of large language models.

While promising, text diffusion techniques have largely not been tested at scale or in multitask transfer learning settings, though see \citet{lin2023text} and \citet{ye2023diffusion} for recent work in this direction.  Furthermore, it remains unclear if these methods demand training new diffusion models from scratch, or if AR models can be efficiently adapted into diffusion models.  We explore these questions empirically in section~\S\ref{sec:experiments}.

One line of previous work shows that non-AR methods benefit from ``AR distillation'' \cite{kim-rush-2016-sequence, gu2018nonautoregressive, saharia-etal-2020-non, gu-kong-2021-fully}---training a non-AR model from scratch on silver data generated via the predictions of an existing AR model.  AR distillation is similar to our AR2Diff adaptation in that both leverage a preexisting AR model. However they differ in that our method initializes the diffusion model directly from an AR checkpoint, and trains on gold data. Given the significant recent investment in training large AR models, we believe that lightweight adaptation of existing checkpoints is a promising direction compared to training non-standard models from scratch.

Recently, \citet{lin2023text} show good results pretraining a text diffusion encoder-decoder model and fine-tuning it on downstream tasks. Like our work, this validates the effectiveness of pretraining text diffusion models at scale.

More recently, building on ``reparameterized discrete diffusion models'' \cite{zheng2023reparameterized}, \citet{ye2023diffusion} show the possibility of converting large AR models (up to 10B parameters) into text diffusion models during task-specific fine-tuning---their ``diffusive adaptation''. This work shares our goal of demonstrating that text diffusion can be practical at scale. Our work differs in (i)~building on SUNDAE as opposed to RDM, (ii)~including diffusion models pretrained from scratch as baselines, (iii)~comparing different architectures and objectives for diffusion pretraining, and (iv)~testing adaptation during pretraining (our AR2Diff$_N$ with $N$\,$>$\,$0$), as opposed to only during fine-tuning (our AR2Diff$_0$).

\section{Evaluation Tasks}
\label{sec:evaluation_tasks}

We experiment with three downstream tasks. First, we use \textbf{WMT14 French-English translation} \cite{bojar-etal-2014-findings}, as machine translation is widely used to evaluate generative models, particularly in work on non-AR models.

Second, we evaluate on the popular \textbf{SQuAD question answering task} \cite{rajpurkar-etal-2016-squad}. As an extractive QA task, this does not require open generation, and most targets are fairly short, often just a few words long. While text diffusion models are unlikely to deliver speed gains on tasks with short outputs (see Section~\S\ref{sec:inference_speed}), we feel it is still important to test for quality on text \emph{understanding} tasks. This can help establish whether pretrained diffusion models can be an effective general foundation for language understanding, and ensures that our findings are interpretable within the literature on transfer learning in NLP\@.

Finally, we evaluate on \textbf{Mostly Basic Python Problems (MBPP)} \cite{austin2021program}, a recent benchmark requiring models to generate full solutions to simple Python programming tasks. This task is fairly open-ended, as there are many working solutions to a given task, depending on choices of algorithm, coding style, variable names, and so on. Compared to open-ended natural language generation, this benchmark has clear and meaningful automatic evaluation metrics, as we can run the generated code and assess whether it passes relevant test cases. When tokenized using the PaLM \cite{chowdhery2022palm} vocabulary we adopt in our experiments, median target length is $59$ tokens, and 90th percentile is $150$ tokens.

\section{Experiments}
\label{sec:experiments}

\subsection{Diffusion implementation}

Our diffusion implementation follows SUNDAE \cite{savinov2022stepunrolled}. More specifically, we use standard Transformer \cite{vaswani2017attention} architectures (either encoder-decoder or decoder-only) as implemented in the T5X \cite{roberts2022t5x} library. As SUNDAE performs discrete diffusion in surface token space, the decoder inputs and outputs are tokens, in line with standard AR models. These implementation choices allow us to reuse existing frameworks for autoregressive LLM training with relatively minor changes. As a result, we can easily experiment with using pretrained AR model checkpoints and adapting these to perform text diffusion.

For training, we use the SUNDAE $L^{(1:2)}$ loss, which incorporates one step of ``unrolled denoising'', encouraging the model to be able to refine its single-step predictions further towards the target. More concretely, for target sequence $x$, we randomly corrupt a random proportion of tokens (sampling from a uniform distribution) to produce $x^c$, which is passed as input to the denoising model to produce logits $l_1$. The ``logits loss'' $L^{(1)}$ is the cross-entropy between $l_1$ and $x$. ``Unrolled logits'' are computed by sampling\footnote{We sample from $l_1$ using temperature $0.0$ (argmax), as opposed to SUNDAE's temperature $1.0$, as we found this performed best in early ablations on WMT14, with temperature in \{\,$0.0$, $0.1$, $1.0$\,\}.} from $l_1$ and passing these tokens back as inputs to the denoising model, producing $l_2$. The ``unrolled logits loss'' $L^{(2)}$ is the cross-entropy between $l_2$ and $x$. For the overall loss, we use $L^{(1)} + L^{(2)}$.

For inference, we follow SUNDAE in using low-temperature sampling ($\tau=0.2$), decoding $N$ samples in parallel (we use $N=8$ by default), and reranking them based on ``model score'': the cross-entropy between the decoder input and output logits on the final step of diffusion. We use $10$ diffusion decoding steps by default; thus on tasks with targets longer than $10$ tokens, our diffusion models use fewer decoding steps than an AR model.\footnote{As AR models can cache and reuse activations from earlier sequence positions for subsequent decoding steps (thanks to the causal attention mask), they use significantly fewer FLOPs per step, when other factors are held constant. We do not present a full picture of the speed vs.~quality tradeoffs of text diffusion models here. Previous work has shown that text diffusion can be competitive on speed and quality, even comparing against AR inference with caching enabled \cite{savinov2022stepunrolled}. We assume here that diffusion in $10$ steps is fast enough to have practical value, and focus on quality.} These choices are ablated in section \S\ref{sec:ablations}.

For simplicity, we forgo SUNDAE's target length prediction module, opting instead to let the model learn to predict sequence length end-to-end through the presence of padding tokens observed during training. As a result, our text diffusion models have no additional parameters beyond those within the Transformer (encoder-)decoder.

\subsection{Selecting objective and architecture}
\label{exp-seq-1}

\begin{figure*}[t]
\centering
\includegraphics[width=0.7\textwidth]{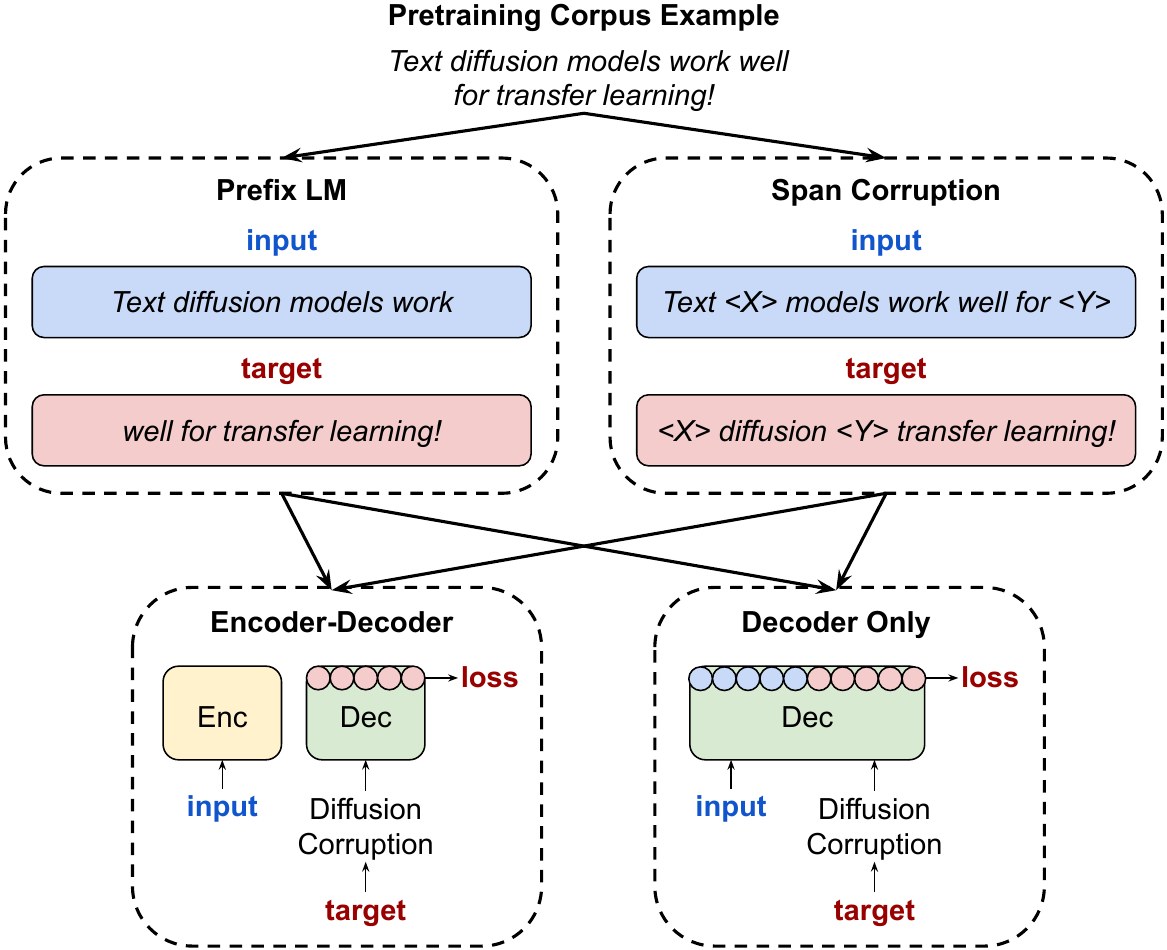}
\caption{Pretraining objectives and model architectures. The <X> and <Y> symbols are unique sentinel tokens denoting masked spans. Note, the ``masking noise'' applied to produce the span corruption input/target is independent from the ``diffusion noise'' which randomly corrupts a subset of target tokens. Loss is only computed over target tokens. In the decoder-only setting, input tokens are frozen when computing the unrolled logits input ($l_2$).}
\label{fig:architectures}
\end{figure*}

\begin{table*}
\centering
\small
\begin{tabular}{llccc}
\toprule
& \textbf{Pretraining} & 
\textbf{WMT14  En-Fr} & \textbf{SQuAD} & \textbf{MBPP}\\
\textbf{Architecture} & \textbf{Objective} & \textbf{(BLEU)} & \textbf{(F1)} & \textbf{(Pass@80 \%)}\\
\midrule
Encoder-Decoder & Prefix LM & 27.6 & 75.8 & 0.0 \\
Decoder-only & Prefix LM & \textbf{29.8} & 77.4 & \textbf{12.2} \\
Encoder-Decoder & Span Corruption & 28.7 & 78.2 & 0.0 \\
Decoder-only & Span Corruption & 29.1 & \textbf{80.6} & 11.1 \\
\bottomrule
\end{tabular}
\caption{\label{tab:exp-seq-1-table}
Diffusion model performance on three tasks across model architecture and pretraining objective. The Decoder-only architecture outperforms Encoder-Decoder across all three tasks, despite using fewer parameters.}
\end{table*}

Previous work on text diffusion has focused on the single-task setting, either training and evaluating on unconditional text generation, or training from scratch on an end task, such as machine translation.\footnote{\citet{ye2023diffusion} adapt pretrained AR models for diffusion across multiple tasks, but do not explore pretraining a general-purpose diffusion model that can be adapted to specific tasks.} In contrast, we aim to evaluate text diffusion in the \emph{transfer learning} setting---pretraining a large model, and adapting it to a range of downstream tasks. As a first step, and to cut down the space of further experiments, we first seek to identify a model architecture and pretraining objective well-suited to text diffusion.

The T5 study on transfer learning for AR text-to-text models \cite{raffel2020t5} recommends using an encoder-decoder architecture and a ``span corruption'' objective---masking multi-token spans in the input, and reconstructing these in the target. By comparison, many subsequent LLMs have converged on a decoder-only architecture with a standard LM objective \cite{brown2020language, chowdhery2022palm}. To establish which setting works best for diffusion, we test all four combinations of architecture (\textbf{encoder-decoder} vs.~\textbf{decoder-only}) and objective (\textbf{span corruption} vs.~\textbf{prefix LM}), as shown in Figure~\ref{fig:architectures}.\footnote{We choose the ``prefix LM'' objective rather than the standard causal LM objective, as it is compatible with the encoder-decoder architecture, and has been shown to outperform causal LM in apples-to-apples comparisons \cite{tay2023ul}.}

We train each model on the same pretraining mixture, consisting of $80$\% multilingual web crawl data from mC4 \cite{xue-etal-2021-mt5} and $20$\% Python code from ``The Stack'' \cite{kocetkov2022stack}. All models use the T5 Base size transformer architecture and pretrain for $1$ million steps on batches of size $128$ and sequence length $1024$. We then fine-tune each model separately on WMT14 En-Fr, SQuAD, and MBPP (producing 12 fine-tuned models total) and evaluate across all tasks. We use a fine-tuning batch size of $128$ and a constant learning rate of $0.001$ across all tasks. We fine-tune $500$K steps for WMT14 En-Fr and $250$K steps for SQuAD, with checkpoints taken every $1{,}000$ steps. For MBPP due to smaller dataset size, we fine-tune for $5{,}000$ steps with checkpoints taken every $50$ steps. In all cases, we terminate fine-tuning if clear evidence of over-fitting is observed. We reuse the $256$K token SentencePiece vocabulary from PaLM \cite{chowdhery2022palm}. Our decoder-only models have roughly $280$M parameters (including embedding parameters), while our encoder-decoder models have roughly $590$M parameters.

\begin{figure*}[t]
\centering
\includegraphics[width=0.7\textwidth]{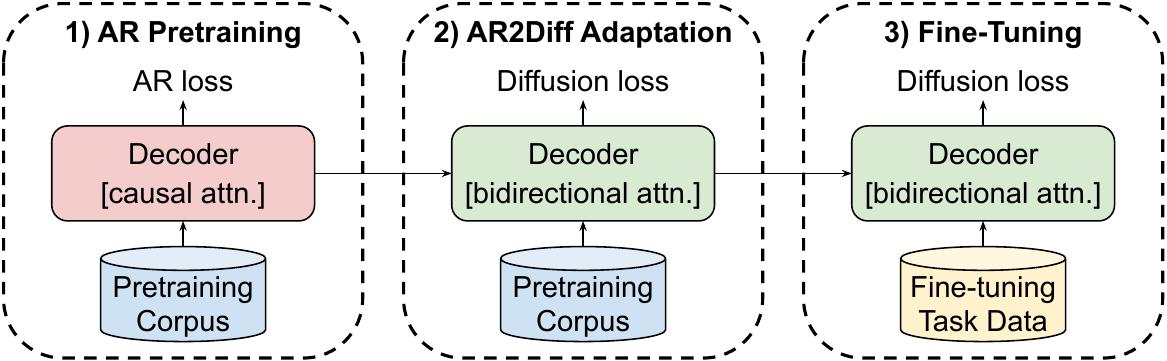}
\caption{Illustration of our AR2Diff method. 1) Pretrain an AR decoder with causal attention. 2) Continue pretraining as a diffusion model with bidirectional attention. 3) Fine-tune as a diffusion model on the end task.}
\label{fig:ar2diff}
\end{figure*}

The results in Table~\ref{tab:exp-seq-1-table} show that our decoder-only models perform the best across all three tasks, despite their lower parameter count. This advantage is especially clear on code synthesis (MBPP), where the encoder-decoder models fail to solve any problem in the test set, even on the permissive ``Pass@80'' metric that samples the model $80$ times and is scored as correct if \emph{any} of these candidates passes. In line with \citet{tay2023ul}, we suspect that pretraining the model to generate longer contiguous spans is a better-matched objective for downstream tasks like MBPP requiring long coherent generation.

Our findings on pretraining objective are less conclusive, with Prefix LM performing the best on WMT and MBPP, while Span Corruption does best on SQuAD\@. With this in mind, we select ``decoder-only + prefix LM'' for our subsequent experiments, as this setup is increasingly standard for LLM training, and does relatively well (best or second-best) across all our tasks.

\subsection{Transfer learning baselines}
\label{sec:transfer_baselines}

We now turn to testing various transfer learning strategies across model scales. As our core baselines, we pretrain both AR and diffusion models at Base ($280$M), Large ($270$M), and XL ($1.7$B) sizes. These all use a decoder-only architecture and prefix LM objective, and train on the same pretraining mixture from the previous section ($80$\% multilingual web pages and $20$\% Python code). As before, we pretrain for $1$M steps, with batch size $128$ and sequence length $1024$. Note, our diffusion models use bidirectional attention to allow modifying all sequence positions in parallel, but are otherwise architecturally identical to their AR counterparts.

For the AR baselines, at inference time, we use greedy decoding for SQuAD, following T5, and use temperature sampling for MBPP, following \citet{austin2021program}. For WMT, we use greedy decoding as opposed to the more commonly used beam search for a fairer comparison, as we did not investigate the use of beam search for diffusion models; see \citet{reid2023diffuser} for work in this direction.

We then fine-tune each of these models separately for each of our three tasks. Results are shown in Table \ref{exp-seq-2-table}, and discussed in section \S\ref{sec:core_results}.

\subsection{AR2Diff: Adapting from AR to diffusion}
\label{sec:ar2diff}

Beyond pure AR and pure diffusion training, we explore ``AR2Diff'' methods for adapting a pretrained AR model into a diffusion model later in training. First, we experiment with simply fine-tuning an AR checkpoint directly using our diffusion training procedure---enabling bidirectional attention, and using the SUNDAE diffusion training loss. We refer to this method as AR2Diff$_{0}$, and use our baseline AR model checkpoint as the starting point for fine-tuning.

We also experiment with pretraining the model for additional steps as a diffusion model \emph{before} fine-tuning, as illustrated in Figure~\ref{fig:ar2diff}. We start with our pretrained AR checkpoint, continue pretraining for an additional $N$ steps using diffusion training, and then fine-tune (still with diffusion) on each evaluation task separately. We refer to this method as AR2Diff$_{N}$.

\subsection{Core results}
\label{sec:core_results}

\begin{table*}
\centering
\small
\begin{tabular}{lcccc}
\toprule
& &\textbf{WMT14 En-Fr} & \textbf{SQuAD} & \textbf{MBPP}\\
\textbf{Method} & \textbf{Size} & \textbf{(BLEU)} & \textbf{(F1)} & \textbf{(Pass@80 \%)}\\
\midrule
Autoregressive & Base & \textbf{33.27} & 68.11 & 5.5 \\
Diffusion  & Base & 29.83 & \textbf{77.41} & \textbf{12.2} \\
AR2Diff$_0$ & Base& 29.62 & 64.77 & 1.1 \\
AR2Diff$_{10,000}$ & Base & 29.41 & 68.12 & 4.4 \\
AR2Diff$_{100,000}$ & Base & 29.92 & 71.87 & 7.7 \\
\midrule
Autoregressive & Large & \textbf{34.92} & 78.43 & \textbf{15.5} \\
Diffusion  & Large & 29.36 & 80.56 & 12.2 \\
AR2Diff$_0$ & Large & 31.14 & 77.82 & 3.3 \\
AR2Diff$_{10,000}$ & Large & 31.97 & 79.62 & 8.8 \\
AR2Diff$_{100,000}$ & Large & 32.20 & \textbf{80.71} & 10.0 \\
\midrule
Autoregressive & XL & \textbf{35.48} & \textbf{84.08} & 15.5 \\
Diffusion  & XL & 29.30 & 82.78 & \textbf{18.8} \\
AR2Diff$_0$ & XL & 32.36 & 80.95 & 6.6 \\
AR2Diff$_{10,000}$ & XL & 32.39 & 80.71 & 11.1 \\
AR2Diff$_{100,000}$ & XL & 32.55 & 83.54 & 15.5 \\
\bottomrule

\end{tabular}
\caption{\label{exp-seq-2-table}
Performance of various models across three tasks and three sizes, comparing: (i) an AR baseline, (ii) a diffusion baseline, and (iii) AR2Diff models that adapt the pretrained AR baseline via diffusion training for $N$ steps before fine-tuning using diffusion, with $N$ $\in$ \{$0$, $10$K, $100$K\}.}
\end{table*}

Results comparing AR2Diff to our autoregressive and diffusion baselines across model sizes are shown in Table \ref{exp-seq-2-table}.

On WMT14 En-Fr, the AR baseline performs the best across model sizes.\footnote{We note our Base AR baseline underperforms ($32.27$ vs.~$37.5$) a similar baseline from \citet{raffel2020t5}, a Base size decoder-only model trained with the same prefix LM objective. This could stem from differences in pretraining data, model architecture, fine-tuning procedure, and/or inference settings (e.g., our use of greedy decoding).\label{footnote:wmt}}  Our observed gap between diffusion and AR is larger than that of \citet{savinov2022stepunrolled}, where SUNDAE text diffusion comes with $1$ BLEU point of an AR baseline.  The difference may be due to our (i) using a transfer learning setting where we pretrain before fine-tuning, (ii) not using SUNDAE's length prediction module, (iii) sampling fewer candidates at inference time ($8$ vs.~$16$). 

Interestingly, while at Base size AR2Diff provides no advantage on WMT, at Large and XL sizes we see AR2Diff delivers a significant gain over the pure diffusion baseline, and this gain increases with the length of adaptation. This suggests that AR2Diff may be valuable not just as a resource-saving method (leveraging AR checkpoints to avoid pretraining diffusion models from scratch), but also as a means of achieving stronger diffusion models through mixed-objective training.

On SQuAD question answering, our diffusion baseline outperforms the AR baseline at Base and Large sizes (Base: $68.1$\,$\rightarrow$\,$77.4$, Large: $78.4$\,$\rightarrow$\,$80.6$), but underperforms at XL size ($84.1$\,$\rightarrow$\,$82.8$).\footnote{As on WMT, these scores are below the results reported by \citet{raffel2020t5} using a similar baseline ($85.4$). See footnote \ref{footnote:wmt}.} While adapting to diffusion only during fine-tuning (AR2Diff$_0$) is ineffective, adapting for $N$ steps before fine-tuning (AR2Diff$_N$) outperforms the AR baseline at most sizes, and improves monotonically with $N$.

On MBPP code synthesis, diffusion outperforms the AR baseline for two out of three model sizes, including the largest XL size ($15.5$\,$\rightarrow$\,$18.8$). As on other tasks, AR2Diff tends to improve with longer adaptation before fine-tuning.

\subsection{Ablations}
\label{sec:ablations}

\begin{table*}
\centering
\small
\begin{tabular}{lcccc}
\toprule
& & & \textbf{SQuAD} & \textbf{MBPP}\\
\textbf{Method} & \textbf{steps} & \textbf{samples} & \textbf{(F1)} & \textbf{(Pass@80 \%)}\\
\midrule
Autoregressive & - & - & 68.11 & 5.5 \\
\midrule
Diffusion  & 5 & 8 & 77.41 & 5.5 \\
Diffusion  & 10 & 8 & 77.41 & 12.2 \\
Diffusion  & 20 & 8 & \textbf{77.72} & \textbf{16.7} \\
\midrule
Diffusion  & 10 & 4 & \textbf{77.51} & 11.1 \\
Diffusion  & 10 & 8 & 77.41 & 12.2 \\
Diffusion  & 10 & 16 & 77.13 & \textbf{13.3} \\
\bottomrule
\end{tabular}
\caption{\label{tab:ablate_steps_samples}
Ablations on diffusion inference hyperparameters \texttt{num\_steps} and \texttt{num\_samples}. Increasing steps and samples leads to clear gains on MBPP, which requires long-form code synthesis, while the effects on SQuAD extractive QA are marginal.}
\end{table*}

Our results so far have performed diffusion inference by running $10$ steps (``\texttt{num\_steps}'') of denoising over $8$ randomly sampled decoder inputs per example (``\texttt{num\_samples}'').  Note, only the output with the highest model score is used for evaluation.  Table~\ref{tab:ablate_steps_samples} shows the results of varying \texttt{num\_steps} $\in$ \{$5$, $10$, $20$\} and \texttt{num\_samples} $\in$ \{$4$, $8$, $16$\}.  On the MBPP code synthesis task, we find that increasing step and samples boosts performance, in line with \citet{savinov2022stepunrolled}. Increasing denoising steps is particularly helpful ($5.5$\,$\rightarrow$\,$16.7$), but at the cost of slower inference. On SQuAD the effect of these parameters is more marginal. More generally, we suspect that additional steps and samples may be helpful on long-form text generation tasks like MBPP that are relatively underspecified (e.g., admitting many correct answers in different styles). By comparison, SQuAD targets are typically short, and are constrained to be spans from the input.

\subsection{Inference speed analysis}
\label{sec:inference_speed}

Diffusion language models have the potential to reduce inference serving costs of long text generation, compared with AR models. Here we show some preliminary results on the inference speed quantitatively. We decode sequences of equal length with AR and diffusion models, and measure corresponding wall-clock times. For diffusion models, we use $10$ diffusion steps as our base case, matching our primary evaluation setup for the WMT, SQuAD and MBPP tasks.

\begin{figure}
\centering
\includegraphics[width=0.93\columnwidth]{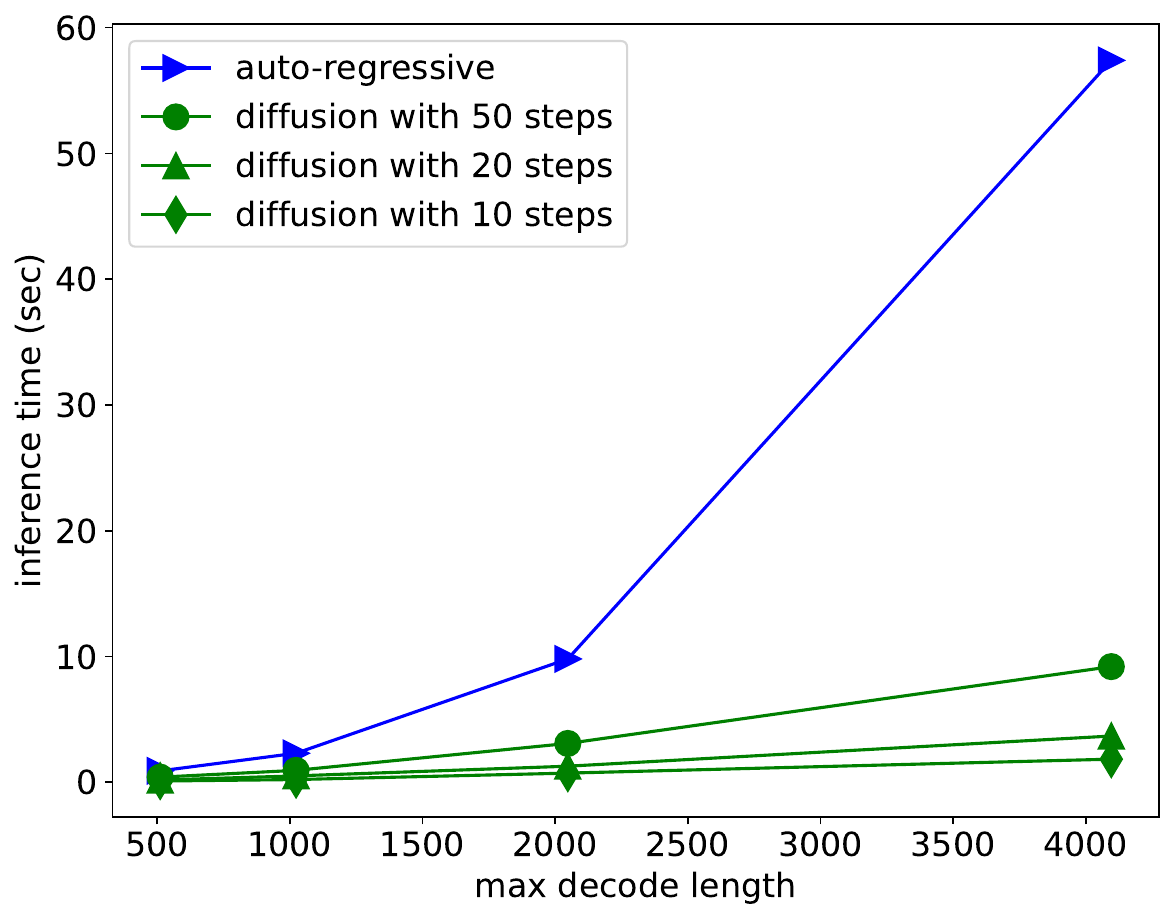}
\caption{By varying the decoding sequence length, we measure inference time of autoregressive decoding vs.~diffusion decoding}
\label{fig:inference_speed_plot}
\end{figure}

We observe an increasing advantage of using diffusion for inference speedup when the generation is long. Figure \ref{fig:inference_speed_plot} shows as the decoding sequence length increases from $500$ tokens (e.g., MBPP task) to $4{,}000$ tokens, the speedup gained by diffusion (using $10$ steps) increases from $10\times$ to $30\times$.

Note that a single AR decoding step ($14$ ms per token generated) is still much faster than a single diffusion step ($179$ ms per denoising step) in our implementation. This is likely due to the diffusion model's lacking the key-value caching widely used to optimize AR inference. Whether caching or other efficiency optimizations can further extend the speed gains of diffusion is an interesting question for future research.

\section*{Acknowledgments}

We are grateful to Jiaxin Shi for helpful comments on an earlier draft.

\bibliography{anthology,custom}
\bibliographystyle{acl_natbib}

\end{document}